\documentclass[10pt,twocolumn,letterpaper]{article}

\usepackage[pagenumbers]{cvpr} 
\usepackage{booktabs,longtable}

\definecolor{darkred}{rgb}{0.7,0.1,0.1}
\definecolor{darkgreen}{rgb}{0.1,0.6,0.1}
\definecolor{cyan}{rgb}{0.7,0.0,0.7}
\definecolor{otherblue}{rgb}{0.1,0.4,0.8}
\definecolor{maroon}{rgb}{0.76,.13,.28}
\definecolor{burntorange}{rgb}{0.81,.33,0}
\definecolor{olive}{RGB}{186, 184, 108}

\usepackage{multirow}

\definecolor{cvprblue}{rgb}{0.21,0.49,0.74}
\usepackage[pagebackref,breaklinks,colorlinks,allcolors=cvprblue]{hyperref}
\usepackage{cuted}
\usepackage{capt-of}

\title{CARLoS: Retrieval via Concise Assessment Representation of LoRAs at Scale}

\author{
    Shahar Sarfaty \quad
    Adi Haviv \quad
    Uri Hacohen \quad
    Niva Elkin-Koren \quad
    Roi Livni \quad
    Amit H. Bermano \\ \\
    {\normalfont \textbf{Tel Aviv University}}
}

\begin{document}

\maketitle
\begin{strip}
  \centering
  \includegraphics[width=\textwidth]{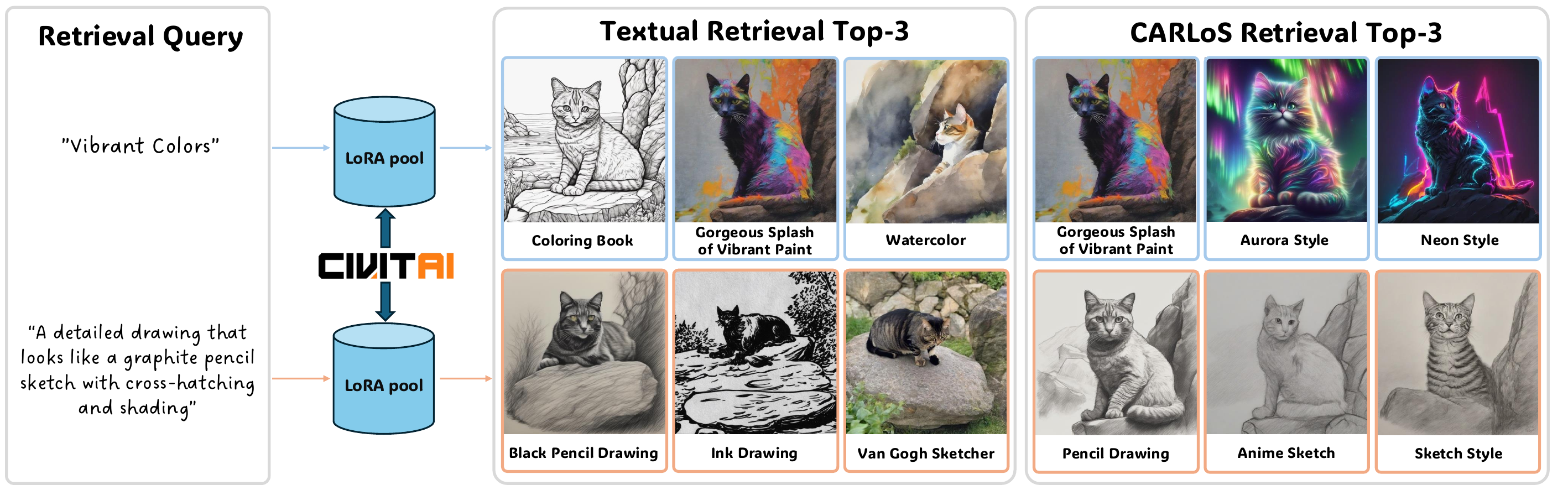}
  \captionsetup{hypcap=false}
  \captionof{figure}{Given a large pool of community generated Low Rank Adapters (LoRAs), our method, CARLoS, concisely represents them according to their influence on generation, and retrieves semantically relevant ones given a retrieval query (left). Since our efficient retrieval (right) is generation based, it finds the LoRAs that are visually similar to the query, outperforming retrieval methods that rely on the name and textual descriptions provided by LoRA creators (middle).}
  \label{fig:teaser}
\end{strip}

\begin{abstract}
The rapid proliferation of generative components, such as LoRAs, has created a vast but unstructured ecosystem. Existing discovery methods depend on unreliable user descriptions or biased popularity metrics, hindering usability. We present CARLoS, a large-scale framework for characterizing LoRAs without requiring additional metadata. Analyzing over 650 LoRAs, we employ them in image generation over a variety of prompts and seeds, as a credible way to assess their behavior. Using CLIP embeddings and their difference to a base-model generation, we concisely define a three-part representation: \textit{Directions}, defining semantic shift; \textit{Strength}, quantifying the significance of the effect; and \textit{Consistency}, quantifying how stable the effect is. Using these representations, we develop an efficient retrieval framework that semantically matches textual queries to relevant LoRAs while filtering overly strong or unstable ones, outperforming textual baselines in automated and human evaluations. While retrieval is our primary focus, the same representation also supports analyses linking Strength and Consistency to legal notions of substantiality and volition, key considerations in copyright, positioning CARLoS as a practical system with broader relevance for LoRA analysis.
\end{abstract}     
\section{Introduction}
\label{sec:intro}

Visual Generation nowadays consists of more than just a single model, but rather a complete ecosystem. Using frameworks such as ComfyUI~\cite{comfyui2023}, generation artists employ dozens of components to refine their generations, including segmentation, personalization, and regeneration of regions. Arguably, the most influential components in such pipelines, except the generative model itself, are Low Rank Adapters (LoRAs)~\cite{hu2021lora}. The open source community has created and published hundreds of thousands of such LoRAs, nudging generation towards various effects, including styles, atmospheres, and specific content (e.g., cat ears). 

This explosion of possibilities leaves the generative artist with a zoo of adapters and very little understanding of their effect. LoRA creators often do not share the data they trained on, leave minimal to no textual description of how their LoRA affects generation, and have no quantifiable means to describe the extent and stability of these effects. In practice, this means that choosing and using the right LoRA requires significant trial and error, with sometimes frustrating and unpredictable results. Similarly, these challenges hinder LoRA creators, users, and especially hosting platforms to screen potentially copyright infringing models.

Previous work that addresses LoRA selection or routing mostly considers the language domain ~\cite{zhao2024loraretriever,muqeeth2024learning, su2025prag, ostapenko2024modular, logo2024}. Recognizing the organizational need in the rapidly evolving and largely unstructured landscape of LoRAs, in the visual domain, recent work typically addresses this task through the textual descriptions, given names, community generated content and other metadata attached to the LoRAs \cite{luo2024stylus,qorbani2025semantic}. These approaches are unreliable predictors of LoRA behavior, or are unavailable. Moreover, this previous work focuses on selection and fusion policies, limiting control and applicability to prompt independent adapter management for expert users, creators, and platforms. 

In this paper, we present CARLoS (\textbf{\underline{C}}oncise \textbf{\underline{A}}ssessment \textbf{\underline{R}}epresentation of \textbf{\underline{Lo}}RAs at \textbf{\underline{S}}cale), a simple and standardized representation for LoRAs, requiring no additional information except for the adapters themselves. As we show, CARLoS efficiently assesses the effect of a given LoRA, aiding in retrieving the correct LoRA for a desired effect, as well as in other questions such as quality assessment and legal attribution. The idea is simple: given a LoRA, we apply it to a variety of prompts and seeds (since existing examples tend to be biased and entangled with other components) and examine the generated images with and without the LoRA’s effect. We encode the generated images using CLIP~\cite{radford2021learning} and summarize the semantic differences between each LoRA-induced image and its vanilla counterpart into three properties - Direction, Strength, and Consistency. Direction is the CLIP-space vector describing the LoRA's semantic shift, while Strength and Consistency are single scalars, describing the extent and stability of the effect. 

Using this simple representation, we find it easy to filter LoRAs that sway too far from the generation prompt or are unpredictable, and to retrieve an adapter from a large set according to a desired effect query with high quality. In addition, we show that this representations can help address legal questions of attribution and protected expression liability, since the central questions of predictable volition and substantiality align well with Strength and Consistency.

To evaluate our approach, we scraped the largest available LoRA collection, of CIVITAI \cite{civitai2022}, and focused on its largest subset, that uses SDXL \cite{podell2023sdxl} as a backbone. For a representative dataset of 650+ usable LoRAs, we generated the prompts, corresponding images, and our extracted CARLoS representations. Compared to four description-driven approaches, we show clear qualitative, quantitative and subjective advantage in retrieving the desired LoRA using CARLoS. Looking forward, we hope CARLoS paves the way to better standardize LoRA community usage, both for the artist and the courts, as well as other generation complementary components such as personalization tokens, IP-adapters~\cite{ye2023ip}, or ControlNets~\cite{zhang2023adding}. Our LoRA dataset, generation prompts, respective image generations and CARLoS representation, queries benchmark, and code will be released upon acceptance. 

\section{Related Work}
\begin{figure*}[t]
    \centering
    \includegraphics[
        width=\textwidth,
    ]{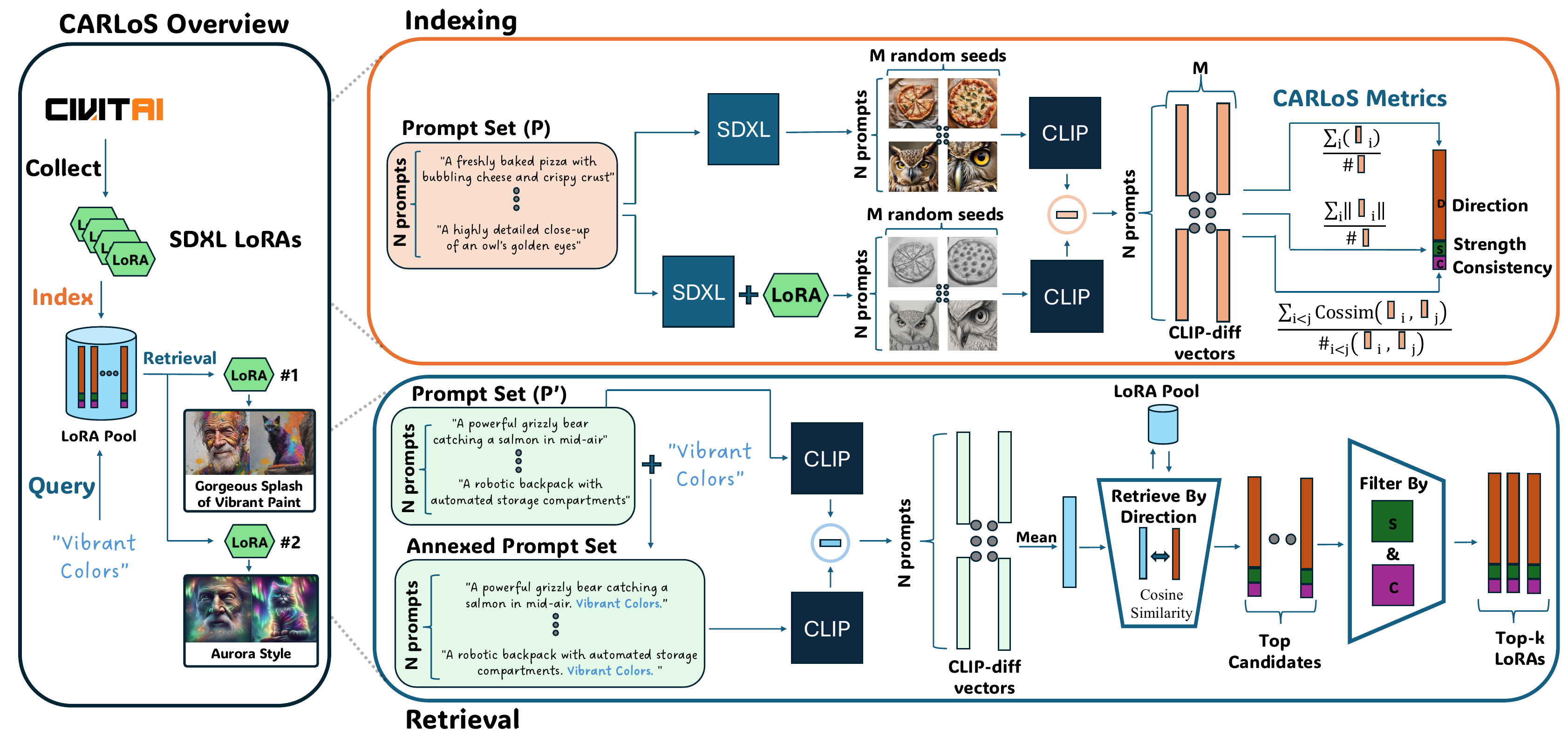}
    \caption{CARLoS framework. Given a set of curated LoRAs operating over the SDXL backbone, we represent each one as a three parts vector, used for efficient retrieval (left). To create our concise representation (top), we generate for each LoRA and the vanilla backbone images using $N=280$ prompts and $M=16$ seeds. We measure the semantic difference between the vanilla generation and the LoRAs in CLIP space (CLIP-diff), and store their average as a representative \textit{Direction} effect, their mean magnitude to represent effect \textit{Strength}, and their variance as a measure for \textit{Consistency}. During retrieval (bottom), we measure the average CLIP space difference between a set of $N$ different prompts with and without the retrieval query appended. We then simply retrieve the LoRAs with the most similar Direction vectors, and filter out LoRAs demonstrating above-threshold Strength and under-threshold Consistency.}
    \label{fig:pipeline} 
\end{figure*}
\paragraph{Adapter Selection and Retrieval.} The growing number of LoRAs has motivated efforts to automatically \textit{select, retrieve,} or \textit{route} adapters for a given prompt.
In the language domain, several methods enable adapter retrieval by measuring similarity between input representations and LoRA-specific signatures. LoRARetriever and PHATGOOSE \cite{zhao2024loraretriever,muqeeth2024learning} train retrieval models using task-specific embeddings, while Parametric-RAG \cite{su2025prag} retrieves adapters based on document similarity. Training-free approaches like Arrow \cite{ostapenko2024modular} and LoGo \cite{logo2024} leverage adapter weight properties for zero-shot selection.
In the visual domain, SemLA retrieves LoRAs for semantic segmentation by aligning input features with the adapters’ training data distributions \cite{qorbani2025semantic}.
Other approaches draw inspiration from Mixture-of-Experts (MoE) frameworks, where a gating network dynamically routes inputs to specialized experts \cite{lepikhin2020gshard,fedus2022switch}.
Finally, most related to our work, systems such as Stylus \cite{luo2024stylus}, LoRAverse
\cite{sonmezer2025loraverse}, AutoLoRA \cite{li2025autolora}, and DiffAgent \cite{zhao2024loraretriever} learn prompt-conditioned selection and fusion policies using metadata, weight embeddings, or gating modules. We do not compare to them because they operate on different LoRA pools (mainly SD 1.5, not SDXL), and are unreleased or require costly retraining. Moreover, these methods are designed for prompt-based composition rather than LoRA search or characterization. In contrast, CARLoS provides a prompt-independent behavioral representation that offers a standardized, metadata-free descriptor complementary to future retrieval and routing systems.

\paragraph{Component Representations} A complementary line of work studies representations of pipeline components, such as prompts, layers, or latent directions, by embedding their semantic effects. 
Previous studies used CLIP-space analysis to quantify semantic shifts induced by edits or tokens, revealing that linear directions in embedding space correspond to interpretable changes in generated images (e.g., color, pose, or texture) \cite{patashnik2021styleclip,abdal2020image2stylegan,kwon2022asyrp}. Subsequent methods such as Prompt-to-Prompt \cite{hertz2022prompt}, StyleAligned \cite{hertz2024stylealigned}, and DiffusionCLIP \cite{kim2022diffusionclip} leveraged this idea to enforce or compare visual consistency across prompts. Other works explore representation learning for generative components, encoding prompts, attention maps, or tokens into compact feature vectors for controllability or attribution \cite{tumanyan2023pnp,liu2023clipscore,parmar2023alignment}. 
We extend this idea to adapters by defining a behavioral embedding that captures each LoRA’s generative effect across prompts—an interpretable, CLIP-based representation requiring no weights or metadata
\label{sec:related}

\newcommand{\myfigheightratio}{0.4}

\section{Method}
\label{sec:method}

Our method, CARLoS (Figure \ref{fig:pipeline}, left), introduces a novel framework for characterizing the effect of text-to-image LoRA adapters.
The method consists of two main stages: (1) curating a corpus of LoRAs and their generations (Section \ref{ssec:dataset}), (2) generating a vector for each adapter representing the overall effect, its Strength, and its Consistency using CLIP-based embeddings and their difference from base-model only generations (Section \ref{ssec:clipdiff}); and (3) application in retrieval (Section \ref{ssec:retrieval}).
\subsection{Dataset Curation}
\label{ssec:dataset}

The first stage of our methodology involves pre-processing for large-scale indexing of LoRA adapters, using a corpus of modules and a comprehensive prompt set.

\paragraph{LoRA Corpus ($\mathcal{L}$).}
We collected SDXL LoRAs, the de facto community standard for the model adaptation stack \cite{frenkel2024blora, thakur2024lcm, hu2021lora, podell2023sdxl}, from the CivitAI platform, which is the largest and most researched public repository for generative models \cite{wei2024exploring, derosa2024civiverse}. The curation involves excluding modules under a minimum age and download count to avoid unstable or transient uploads, Not-Safe-For-Work (NSFW) modules, and corrupted files, which account for a significant portion of potential candidates. After a filtering and validation process, our final corpus $\mathcal{L}$ comprises \textbf{656 valid LoRAs}.

\paragraph{Prompt Set ($\mathcal{P}$).}
We curate a set of English prompts $\mathcal{P}$, crafted using an LLM under human guidance. To avoid user-generated bias and address a comprehensive span of common usage patterns as observed on CivitAI, our guidance focused on $K(=10)$ distinct semantic categories (e.g., ``portraits'', ``animals'', ``fantasy''), semantic variety and constraints on prompt length. Our final representative prompt set contains $N=280$ prompts. 
\newline

Additional implementation details of the LoRAs filtering and prompt set construction processes are provided in the supplementary materials for reproducibility.

\subsection{Adapter Indexing}
\label{ssec:clipdiff}

The core of our analysis relies on isolating the precise generative effect of each LoRA. 
We fix a set of $M$ random seeds $\mathcal{S}$. For each LoRA $l\in\mathcal{L}$, prompt $p\in\mathcal{P}$, and seed $s\in\mathcal{S}$ we generate paired images under the same hyperparameters: $x^{(0)}_{p,s}$ using only the vanilla base model, and $x^{(l)}_{p,s}$ with the LoRA modification. Then, to quantify the LoRA's effect we use a pre-trained CLIP image encoder \cite{radford2021learning} and project both images into the joint text-vision embedding space. Let $v(\cdot)\in\mathbb{R}^d$ be the CLIP image encoder. This \textit{CLIP-diff} vector collection is $\mathcal{V}$ is:
\begin{equation}
\label{eq:clipdiff}
\mathcal{V}=\{v\!\big(x^{(l)}_{p,s}\big) \;-\; v\!\big(x^{(0)}_{p,s}\big)\}_{p\in\mathcal{P},s\in\mathcal{S}}.
\end{equation}

This one-time, resource-intensive generation and embedding process yields a corpus of 512-dimensional CLIP-diff vectors, representing the pure semantic and stylistic shift introduced by each LoRA for a specific prompt and seed.

Using the CLIP-diff corpus $\mathcal{V}$, we quantify each LoRA's effect via three novel metrics. We denote the set of all CLIP-diff vectors for a given LoRA $l$ as $V^l$, and compute the final representation consisting of the three parts: semantic Direction, effect Strength and effect Consistency. 

\paragraph{Semantic Direction.} The average CLIP-diff of LoRA $l$:
\begin{equation}
    \text{SD}(l) = \frac{1}{|V^l|}\sum_{v \in V^l} v
\end{equation}
This 512-dimensional vector represents the average semantic direction, or central tendency, of the LoRA's effect in CLIP space. It serves as a single-vector signature of the LoRA's typical stylistic and semantic impact, and is the core component of our retrieval method.

\paragraph{Strength.} The average LoRA's CLIP-diff vectors norm:
\begin{equation}
    \text{Str}(l) = \frac{1}{|V^l|}\sum_{v \in V^l} \left\lVert v \right\rVert_2
\end{equation}
This scalar metric quantifies the average magnitude of the LoRA's effect. It indicates how significantly the LoRA alters the base SDXL generation, regardless of the specific semantic direction. Although we find that the “LoRA Scale” hyperparameter influences Strength, we fix it (to 1) across all adapters. We find the connection between Strength and Scale is non-trivial (see supplementary materials) across different LoRAs, and leave the computationally intensive investigation of optimal scale usage for each adapter as future work. 

\paragraph{Consistency.} The average pair-wise cosine similarity of the LoRA's CLIP-diff vectors:
\begin{equation}
    \text{Cons}(l) = \frac{1}{\binom{|V^l|}{2}}\sum_{v_i, v_j \in V^l, i < j} \frac{ v_i \cdot v_j}{\left\lVert v_i \right\rVert_2 \left\lVert v_j \right\rVert_2}
\end{equation}
This metric measures the internal coherence of the LoRA's effect. A high consistency score (approaching 1) indicates a predictable and specific semantic shift, regardless of the prompt or seed. Conversely, a low score suggests the LoRA's effect is highly variable, unpredictable, or chaotic, invalidating the confidence of the calculated average direction, effectively hindering downstream tasks. 

\subsection{LoRA Retrieval}
\label{ssec:retrieval}

The primary application of the LoRA representations produced by CARLoS is a novel LoRA retrieval pipeline. (Figure \ref{fig:pipeline} bottom right). Unlike methods that rely on textual or training set related metadata, our method retrieves LoRAs based purely on their generative effect. 

Given a text query $q$ (e.g., "vibrant colors"), its semantic effect can be modeled similarly to $SD$ as a differential vector in CLIP space. Hence, we define an additional, large prompt corpus $\mathcal{P}'$, comparable in size and categorical scope to our generation corpus $\mathcal{P}$. 
$\mathcal{P}'$ was built independently of $\mathcal{P}$ to avoid information leakage.
Let $u(\cdot)\in\mathbb{R}^d$ be the CLIP text encoder. 
For $p'\in\mathcal{P}'$ we define the suffixed prompt $p'\oplus q$, where $\oplus$ denotes concatenation, and the textual diff
\[
\delta_q(p') \;=\; u(p'\oplus q) - u(p').
\]
The \emph{reciprocal textual CLIP-diff} is then
\begin{equation}
\label{eq:query_vec}
\bar{\delta}_q \;=\; \frac{1}{|\mathcal{P}'|} \sum_{p'\in\mathcal{P}'} \delta_q(p').
\end{equation}

Overall, the retrieval process operates in two stages: ranking candidate LoRAs and filtering them based on quality and stability.

First, we perform the primary retrieval. We \textbf{rank} every LoRA $l \in \mathcal{L}$ by computing the cosine similarity between the query vector $\bar{\delta}_q$ and the LoRA's Semantic Direction  $\text{SD}(l)$.

Then, we apply a concise \textbf{filtering} stage to ensure the quality and usability of the results. From the ranked list, we remove LoRAs that are too strong LoRAs (shifting the generation toward specific content, severely hindering prompt adherence), or are unpredictable, with low coherence, where the adapter effect is unreliable. This is done by applying thresholds for both our Strength and Consistency metrics. The final candidate set $\mathcal{L}'$ contains only the LoRAs $l$ from the ranked list that satisfy:
    $$ \mathcal{L}' = \{ l \in \mathcal{L} \mid \min_k(\text{SD}(l) \cdot \bar{\delta}_q),\text{Str}(l) < \tau_{s}, \text{Cons}(l) > \tau_c \} $$
\section{Experiments}
\label{sec:experiments}

To validate the effectiveness of our CARLoS framework, we conduct a series of experiments focused on the downstream task of LoRA retrieval. We demonstrate that our method, which relies on the quantified generative effect of LoRAs, significantly outperforms traditional text retrieval methods based on user-provided metadata.

\subsection{Implementation Details}
\label{sec:implementation}

All images were generated using the Stable Diffusion XL 1.0 base model \cite{podell2023sdxl}. 
For visual and textual CLIP \cite{radford2021learning} embeddings, we use the ViT-B/32 variant.
We generated the images with the default hyper-parameters (e.g. CFG $=7.5$, Euler Scheduler, Lora Scale $=1$). We set the Strength threshold $\tau_s=9.8$ and Consistency threshold $\tau_c=0.041$ in all experiments. The initial indexing process involved generating $(|\mathcal{L}|+1) \times |\mathcal{P}| \times M =  {\sim3M}$ images (for $|\mathcal{L}|=656$ LoRAs, $|\mathcal{P}|=280$ prompts, and  $M=16$ seeds), taking approximately 7 NVIDIA A6000 GPU-hours per LoRA. Once indexed, the retrieval process is highly efficient. Computing a new textual query vector $\overline{\delta}_{q}$ (Eq.~\ref{eq:query_vec}) takes approximately 5 seconds on a single NVIDIA A5000 GPU. Ranking against the pre-computed $SD(l)$ signatures of all 656 LoRAs is near-instantaneous (0.09 seconds). See supplementary materials for additional details. 

\subsection{Experimental Setup}
\label{sec:exp_setup}

\paragraph{Baselines:} We compare CARLoS against four strong, multilingual, text-embedding models \cite{enevoldsen2025mmteb} used for retrieval. These baselines operate on the user-provided textual metadata (concatenated names and descriptions) associated with each LoRA in our corpus. This metadata is often sparse, subjective, and written in various languages, posing a challenge that modern multilingual models are designed to handle. The baselines are Qwen3 \cite{yang2025qwen3}, E5 (Multilingual-E5-instruct) \cite{wang2024multilingual}, BGE (BGE-M3-reranker) \cite{multi2024m3} and GTE (mGTE-reranker) \cite{zhang2024mgte}.

\paragraph{Evaluation Protocol:} To create a robust and comprehensive benchmark, we generated a set of over 700 representative text queries using a combination of large language models (GPT \cite{achiam2023gpt}, Grok \cite{grok-tool}, and Gemini \cite{comanici2025gemini}), that cover  common artistic and conceptual searches. Our quantitative evaluation uses Vision-Language Models (VLMs) as evaluators. For each query, we retrieve top-k LoRAs and generate images for them using a small fixed prompt set. We then evaluate the relevance between the generated images and the concatenated query text using four state-of-the-art VLMs and aesthetic models: Qwen2.5-VL \cite{bai2025qwen2}, SigLIP2 \cite{tschannen2025siglip}, ImageReward (IR) \cite{xu2023imagereward}, and HPS v2 (HPS) \cite{wu2023human}.

To make scores comparable across the different evaluators (which have different and sometimes negative value ranges), we first linearly normalize the scores for each evaluator across all queries and retrieval methods to the [0, 1] range, where 0 maps to the globally lowest score and 1 maps to the highest. We then average the scores across all queries and top-k retrieved LoRAs. This provides a single, robust score for each retrieval method. While our main Table~\ref{tab:quantitative} shows the top-3 results for brevity, similar performance improvements were observed for $k=1$ through $k=7$. See supplementary material for the comprehensive set of results.

\subsection{Qualitative Evaluation}
\label{sec:qualitative}
\begin{figure}[t]
    \centering
    \includegraphics[
        width=\columnwidth,
    ]{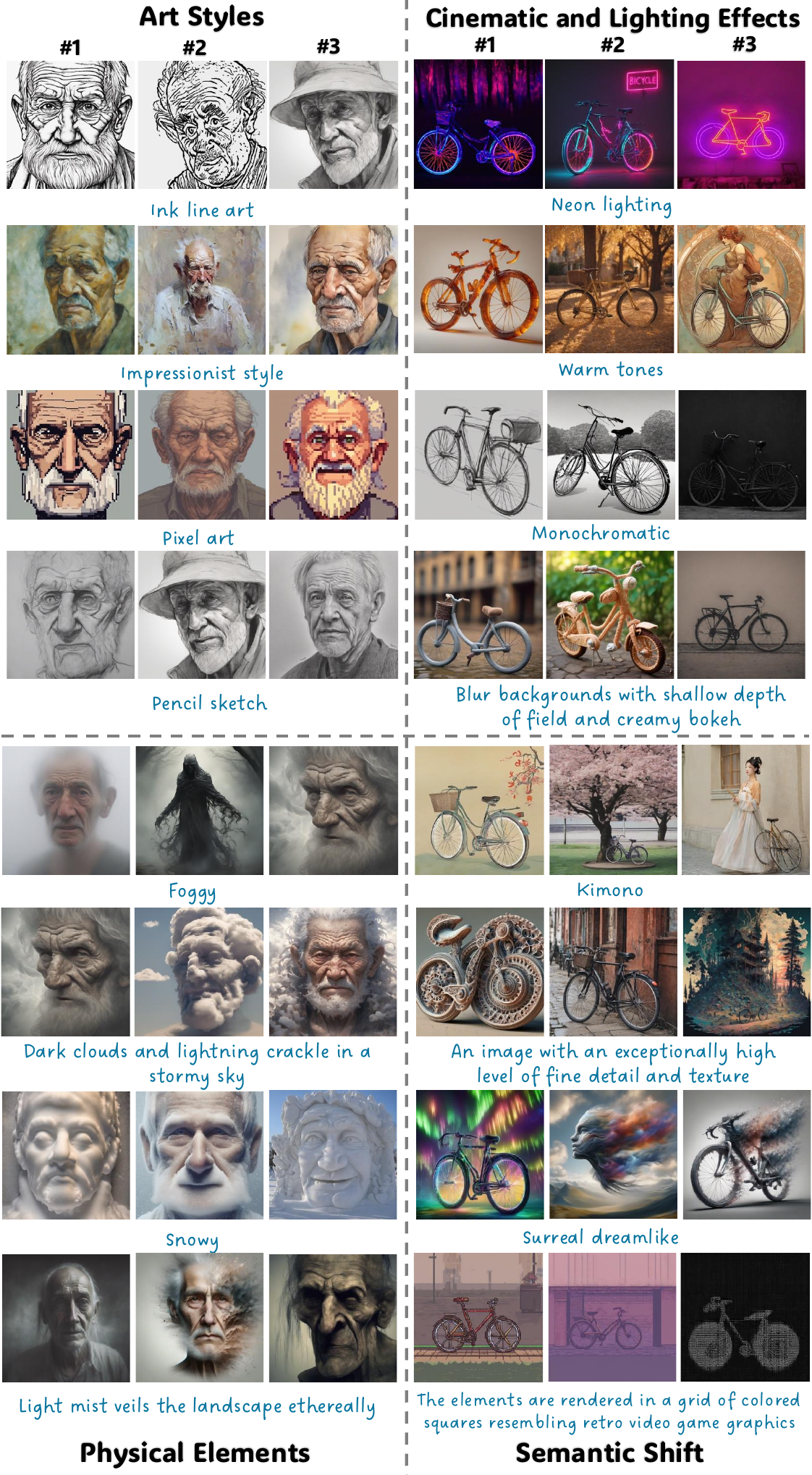}
    \caption{Qualitative retrieval results for CARLoS. Various query modifications are presented, depicting different effect types. The vanilla backbone generated image is on the top left, and its LoRA-modified counterparts are depicted for the top-3 retrieved LoRAs for each query below. Zoomed in viewing recommended. }
    \label{fig:gallery} 
\end{figure}

\begin{figure*}[t]
    \centering
    \includegraphics[
        width=\textwidth,
    ]{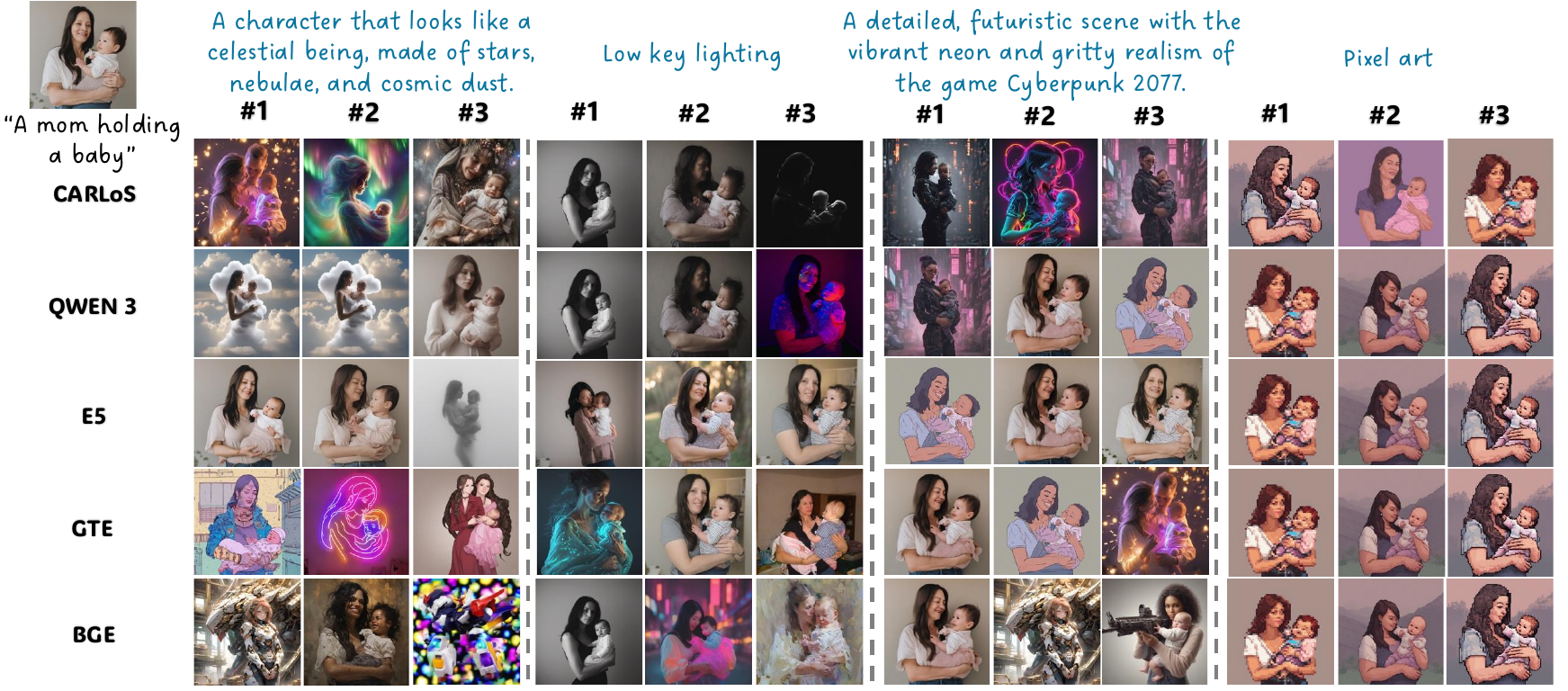}
    \caption{Qualitative comparisons of textual description-based retrieval (bottom rows) to CARLoS (top row). While some effects are sufficiently described in text (e.g., Pixel art) and are therefore retrieved well, more elaborate queries, (such as celestial beings, or futuristic games) are not described well, resorting textual-based retrieval to similar wording as opposed to effects (e.g., clouds, cartoons) }
    \label{fig:qualitative_comparison_to_text_retrieval} 
\end{figure*}

Figure~\ref{fig:gallery} showcases the quality of our retrieval. For diverse queries, our method successfully ranks relevant LoRAs, demonstrating robustness and utility. In particular, for highly abstract (e.g., ''Surreal dreamlike'') or esoteric (e.g. ''Kimono''), which have no exact match, our method still retrieves visually relevant results. We also see visual relevance for LoRAs with textually described unrelated intent (e.g., highly detailed "epic land" LoRA). Furthermore, the figure demonstrates the diversity of retrieved LoRAs across all prompts, which is also supported empirically (see Supplementary Material for quantitative analysis).

\paragraph{Comparative Results:} In addition to the examples shown in Figure~\ref{fig:teaser}, Figure~\ref{fig:qualitative_comparison_to_text_retrieval} provides a qualitative comparison between CARLoS and the four textual baselines. While in simple cases (e.g. ``pixel art'') textual retrievers are comparably effective, they often fail, latching onto irrelevant or confusing keywords in the descriptions or names (e.g. ``Coloring Book'' in Figure~\ref{fig:teaser}), or including unfiltered LoRAs which are too strong (e.g. BGE's "celestial being" result \#3, Figure~\ref{fig:qualitative_comparison_to_text_retrieval}) or inconsistent (e.g., QWEN's "futuristic" result \#2, Figure~\ref{fig:qualitative_comparison_to_text_retrieval}). In contrast, CARLoS consistently retrieves LoRAs that accurately reflect the query's semantic and stylistic intent.

\subsection{Quantitative Evaluation}
\label{sec:quantitative}
As shown in Table~\ref{tab:quantitative}, our CARLoS method consistently outperforms the textual retrieval baselines across all four semantics and aesthetics evaluators. This quantitative result validates our core hypothesis: retrieving LoRAs based on their actual, measured generative effect is significantly more reliable and accurate than relying on biased, inconsistent, and often minimal user-provided text descriptions.
\newcommand{\impr}[1]{\textcolor{green!60!black}{\small\bfseries(+#1\%\ $\uparrow$)}}
\begin{table}[ht]
\centering
\caption{\textbf{Retrieval Performance Evaluated by Different VLMs.} Scores indicate the quality of retrieved top-3 LoRAs as judged by state-of-the-art Vision-Language Models. CARLoS consistently yields results preferred by all evaluators. The scores are normalized in min-max manner across all queries and retrievers.}
\label{tab:quantitative}
\resizebox{\columnwidth}{!}{
\begin{tabular}{@{}p{3.5cm}ccccc@{}}
\toprule
\textbf{Retriever} & \textbf{SigLIP2} & \textbf{Qwen2.5} & \textbf{IR} & \textbf{HPS} \\
\midrule
E5 & 0.289 & 0.480 & 0.449 & 0.565\\
GTE & 0.258 & 0.461 & 0.439 & 0.556\\
BGE & 0.199 & 0.429 & 0.387 & 0.543\\
Qwen3 & \underline{0.307} & \underline{0.495} & \underline{0.491} & \underline{0.590} \\
\textbf{CARLoS} & \textbf{0.350} & \textbf{0.532} & \textbf{0.505} & \textbf{0.596} \\
\bottomrule
\end{tabular}
}
\end{table}

\subsection{Subjective User Study}
\label{sec:user_study}

To validate our quantitative VLM-based evaluations with human judgment, we conducted a user study. We tasked 36 human participants with a series of A/B comparison questions (see Supplementary Material for questionnaire screen shot and additional details). In each of the approximately 100 unique questions, participants were shown two sets of images: one generated using the top-3 LoRAs retrieved by CARLoS, and the other using the top-3 LoRAs from one of the four textual baseline methods for the same query. Each question, answered by at least 6 different individuals, asked participants to choose the winning set based on \textit{Images Quality}, \textit{Relevance to the LoRA Query}, and their \textit{Overall Preference}.

\begin{figure}[ht]
    \centering
    \includegraphics[width=\columnwidth]
    {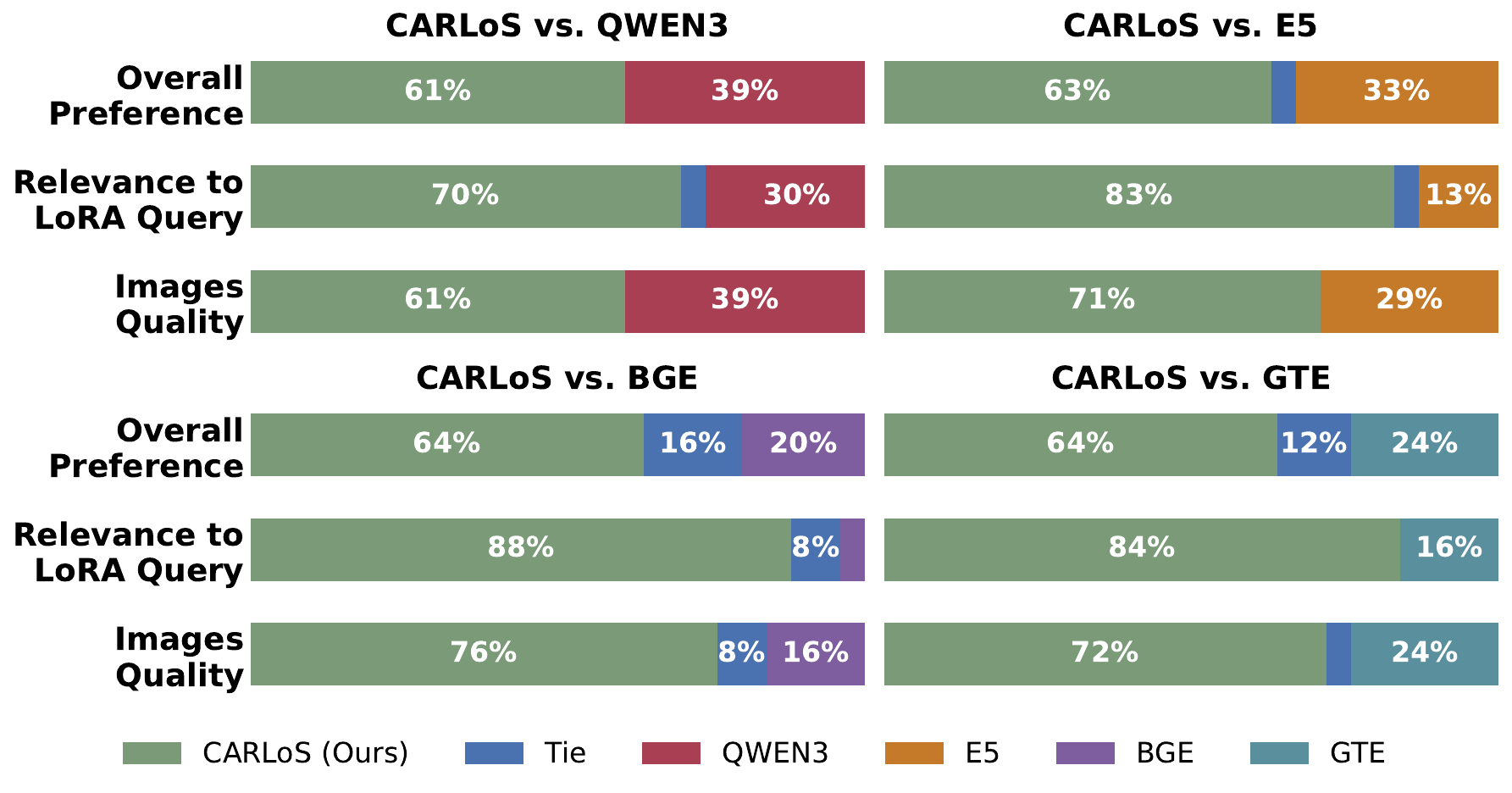}
    \caption{Aggregated results of our subjective user study. Participants compared CARLoS against four strong textual retrieval baselines (QWEN3, E5, BGE, GTE) across three criteria. CARLoS was consistently preferred in all categories.}
    \label{fig:user_study}
\end{figure}

The findings demonstrate a clear and consistent human preference for the LoRAs retrieved by CARLoS across all three evaluation metrics for all four textual retrieval baselines (see Figure~\ref{fig:user_study}). This subjective validation strongly corroborates our quantitative findings, confirming that CARLoS identifies LoRAs are more relevant as well as appealing.

\newcommand{\decline}[1]{\textcolor{red!60!black}{\small\bfseries(-#1\%\ $\downarrow$)}}
\newcommand{\improve}[1]{\textcolor{green!60!black}{\small\bfseries(+#1\%\ $\uparrow$)}}
\begin{table}[t]
\centering
\caption{\textbf{Ablations.} Our full method is compared against method variants, evaluating the contribution of design choices. For comparability, the scores are mapped using the normalization calculated for the retrievers performance comparison.}
\label{tab:ablations}
\resizebox{\columnwidth}{!}{
\begin{tabular}{@{}lcccc@{}}
\toprule
\textbf{Variant} & \textbf{SigLIP2} & \textbf{Qwen2.5} & \textbf{IR} & \textbf{HPS} \\
\midrule
\textbf{Full} & \textbf{0.350} & \textbf{0.532} & \textbf{0.505} & \underline{0.596} \\
No Strength Filtering & 0.335 & 0.525 & 0.495 & 0.596 \\
No Consistency Filtering & 0.342 & 0.529 & \underline{0.501} & \textbf{0.599} \\
No Filtering & 0.335 & 0.525 & 0.495 & 0.596 \\
Query as Prefix & 0.338 & 0.523 & 0.488 & 0.589 \\
Query as Prefix \& Suffix & \underline{0.344} & \underline{0.530} & 0.495 & 0.592 \\
Only Query & 0.328 & 0.511 & 0.426 & 0.538 \\
\bottomrule
\end{tabular}
}
\end{table}
\subsection{Ablation Study} We conducted an ablation study, detailed in Table~\ref{tab:ablations}, to analyze the contribution of the key components of our retrieval system:
Removing Strength and Consistency \textbf{filtering}, both separately and jointly, consistently degrades performance.
We also tested different methods for creating the \textbf{textual query}. Our method annexes the query as a suffix to prompts, outperforming variants of prefix, combined prefix and suffix, or using only the query. This confirms that our reciprocal approach effectively captures the semantic intent.

\subsection{Analysis}
\label{sec:lora_analysis}

\begin{figure}[t]
    \centering
    \includegraphics[
        width=\columnwidth
    ]{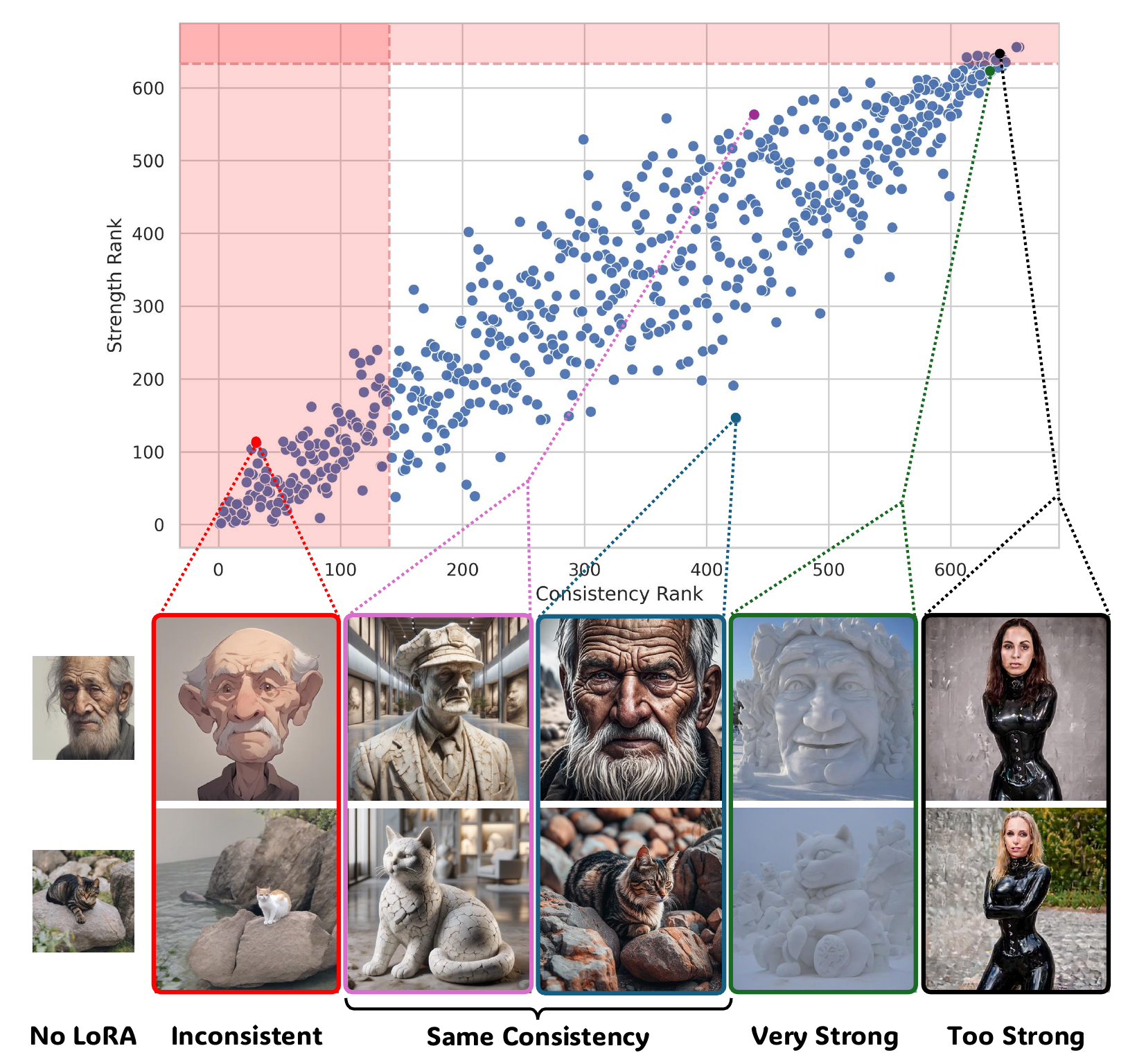}
    \caption{Top: Our LoRA dataset, in Consistency Rank vs. Strength Rank distribution. Too strong and too inconsistent LoRAs (red regions) are filtered out. Bottom: Example generations for two prompts for different strength and consistency LoRAs.}
    \label{fig:consistency-strength-scatter} 
\end{figure}

A core contribution of our work is the ability to characterize the LoRA ecosystem. Figure~\ref{fig:consistency-strength-scatter} plots our entire corpus of 656 LoRAs on a 2D scatter plot, with Consistency Ranking on the x-axis and Strength Ranking on the y-axis. This visualization reveals the distribution of LoRA behaviors.

The qualitative examples at the bottom of Figure~\ref{fig:consistency-strength-scatter} illustrate:
an \textbf{inconsistent} LoRA, which produces different effects for different prompts and seeds (see Figure~\ref{fig:law_figure} top-right);
LoRAs that apply a consistent effect across different prompts but change the semantic content in \textbf{different strengths}. For example, the lighting change in the blue frame is as coherent, but not as strong as the subject material and scene alternation in the purple frame;
\textbf{Strong} LoRAs. This highlights the necessity of our Strength filter. A ``Very Strong'' LoRA may be desirable, but a ``Too Strong'' LoRA completely overwrites the base prompt (e.g., ignoring ``the face of an old man'' to produce a specific character), making it unsuitable for general use. The red shaded region in the plot indicates LoRAs that are filtered out during retrieval to ensure high-quality, prompt-adherent results.

Furthermore, although Strength is influenced by the LoRA Scale hyperparameter, we observe that the functional connection between Strength and scale is not strictly linear, and behaves differently for weak and strong LoRAs. Thus, this  hyperparameter is not a reliable predictor for Strength. See supplementary for a detailed analysis on the influence of the LoRA scale on the Strength.
\section{Legal Application}
\label{sec:legal}

Researchers have recently demonstrated that modern diffusion models, both foundation and LoRA-based, can memorize and reproduce protected expression, thereby exposing users, model designers, and platforms hosting these models to potential copyright liability \cite{he2024fantastic,chiba2025tackling,lee2024talkin}. For example, the Hangzhou Internet Court in China recently held a marketplace platform liable for failing to remove a LoRA that reproduced the protected features of the well-known character Ultraman \cite{meuwissen2025ai_copyright,eu_iphelpdesk_ultraman2025}. Our proposed CARLoS metrics offer a robust and quantifiable framework to aid in addressing these emerging legal considerations, providing an assessment of a LoRA's influence on generation. 

We note that not all instances of memorization and reproduction by LoRAs will automatically trigger liability. Under U.S. copyright law, liability for direct copyright infringement requires a volitional act that results in more than de minims copying of protected expression that is substantially similar to the original copyrighted work by \citet{denicola2015volition}. Thus, model designers, users, and hosting platforms will not be directly liable for negligible  memorization and reproduction of protected expression \cite{ringgold_v_bet_1997}. Nor will they be liable for unexpected memorization and reproduction that they could not reasonably foresee or control \cite{goranin2024deep,chatterjee7minds}.

\begin{figure}[t]
    \centering
    \includegraphics[
        width=0.90\columnwidth
    ]{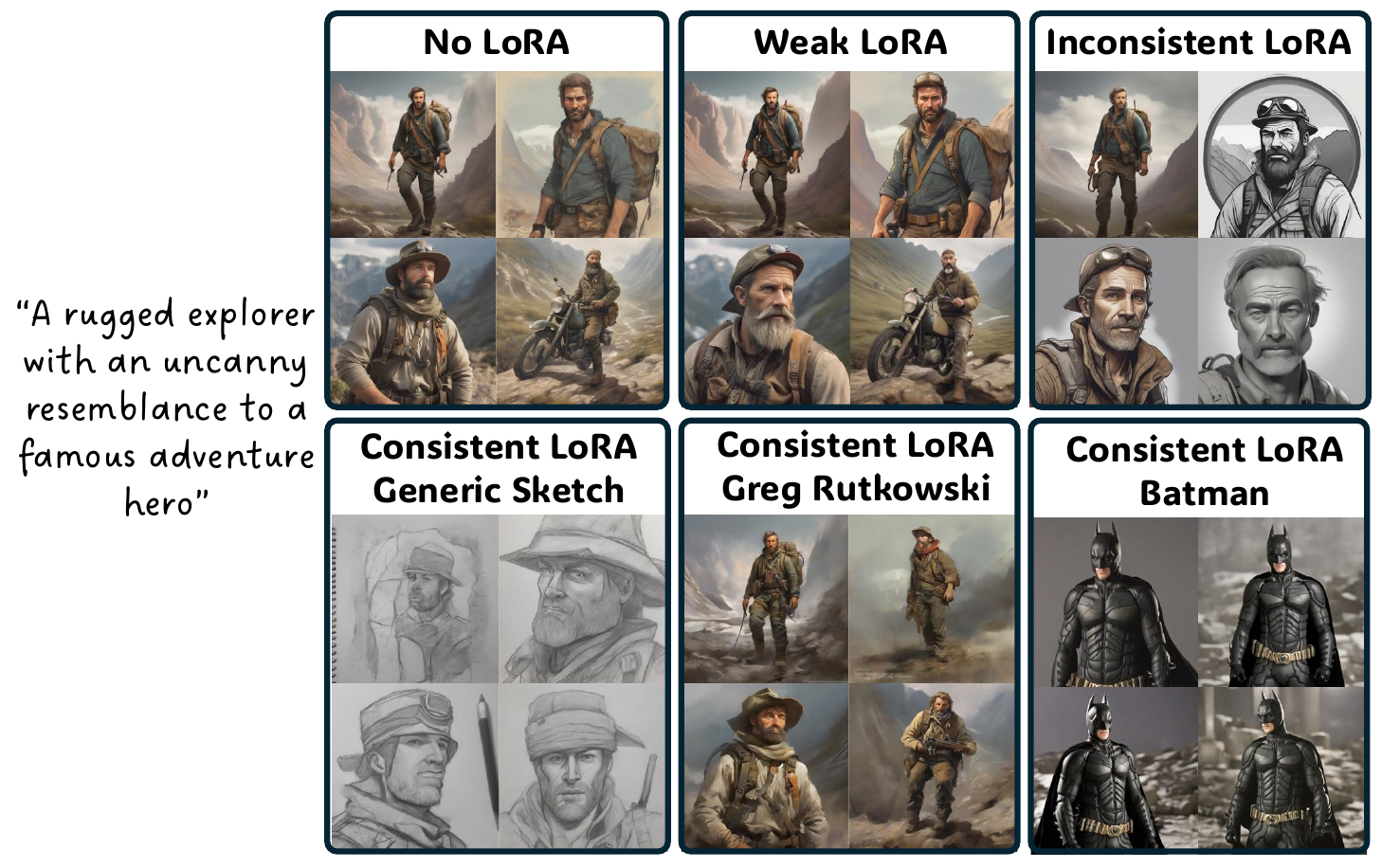}
    \caption{Legal considerations. LoRAs expose users to liability and rights. Weak or inconsistent LoRAs (top), are unlikely to impose infringement or authorship. Strong consistent LoRAs may infringe depending on protected elements replication, or distinct styles (bottom-right,middle), but not necessarily (bottom-left).
    }
    \label{fig:law_figure} 
\end{figure}

The two measures we propose here -- Strength and Consistency -- serve as proxies for these legal elements of substantiality and volition, respectively. Weak LoRAs (see Figure \ref{fig:law_figure}, top–middle) are unlikely to reproduce substantial enough protected expression to create direct liability \cite{ringgold_v_bet_1997}. Similarly, inconsistent LoRAs, even if strong (see Figure \ref{fig:law_figure}, top–right), are unlikely to give rise to liability since their creators and users cannot reasonably predict when or whether the model will reproduce protectable material, meaning the element of volitional control is absent \cite{vht_v_zillow_2019,burcher1997religious}.

LoRAs that are both strong and consistent may infringe when they replicate protected elements such as the recognizable features of a copyrighted character (Figure \ref{fig:law_figure}, bottom–right), as in the Hangzhou case. LoRAs may also consistently reproduce the distinctive style of specific visual artists (Figure \ref{fig:law_figure}, bottom–middle). While distinctive style is not generally protected under copyright law \cite{neilsen2024protection,sobel2024elements,jurcys2023protecting}, it remains ethically and commercially contested, prompting many hosting platforms to voluntary self-regulate \cite{bastian2025openai_artist}.

Accordingly, LoRAs that are both weak and inconsistent will never meet the threshold for direct copyright liability, while strong and consistent ones might. These metrics therefore offer an intuitive, scalable tool for LoRA creators, users, and especially hosting platforms to screen potentially infringing models and mitigate exposure to liability. Importantly, even using a LoRA that is both strong and consistent would not necessarily amount to direct copyright infringement (Figure \ref{fig:law_figure}, bottom left). The key question -- whether the model is reproducing protected expression -- is beyond the scope of our proposed metrics, and will require separate methods to assess \cite{hacohen2023copyright,chiba2025tackling}.

As a final note, the converse of copyright liability is attribution. As the U.S. Copyright Office recently clarified, model designers and users typically lack protectable interests in AI-generated outputs because “there is no guarantee that a particular prompt will generate any particular visual output” \cite{usco2023_zarya}. However, where a strong and consistent LoRA reproduces original expression first introduced by its designer, the designer and subsequent users may attempt to claim copyright protection as derivative works, subject to the designer’s licensing terms \cite{uscode2018,rachum2018copyright}.

\section{Discussion}
\label{sec:discussion}

We presented \textbf{CARLoS}, a large-scale framework that quantitatively characterizes the generative effect of text-to-image LoRAs through their measured semantic impact in CLIP space, based solely on their visual generations. Our concise metrics (Semantic Direction, Strength, and Consistency), enables robust semantic adapter retrieval.

Moreover, the proposed metrics provide interpretable proxies for legal notions of substantiality and volitional control, illustrating their potential beyond retrieval tasks.

\paragraph{Limitations and Future Work.}
Our method inherits the constraints of the visual encoder CLIP, known to be weaker in spatial composition, fine-grained texture, and multi-modal biases. Examining the effects of newer VLMs is warranted.
In addition, due to computational restrictions, all generations were performed with the same hyperparameters. Specifically,  a LoRA scale of $1$ is used. Strength and scale are different in nature, however investigating more diverse values is an important avenue. Similarly, the extension to other backbones (e.g., SD 3, FLUX) \cite{esser2024scaling,duss-taval2024flux} and adapter types (e.g., ControlNets, IP-Adapters) may reveal different behavioral distributions.

Finally, the one-time LoRA indexing process requires approximately 7 GPU hours per adapter. This  poses scalability challenges for high-throughput platforms.

\paragraph{}
In conclusion, we envision CARLoS fostering a standardized and transparent ecosystem for managing community-driven generative components. The representation introduced could facilitate interpretability, usability, and help the generative ecosystem grow. 

{
    \small
    \bibliographystyle{ieeenat_fullname}
    \bibliography{main}
}

\clearpage
\appendix
\setcounter{page}{1}
\maketitlesupplementary

\begin{figure*}[ht]
    \centering
    \includegraphics[width=0.75\textwidth]{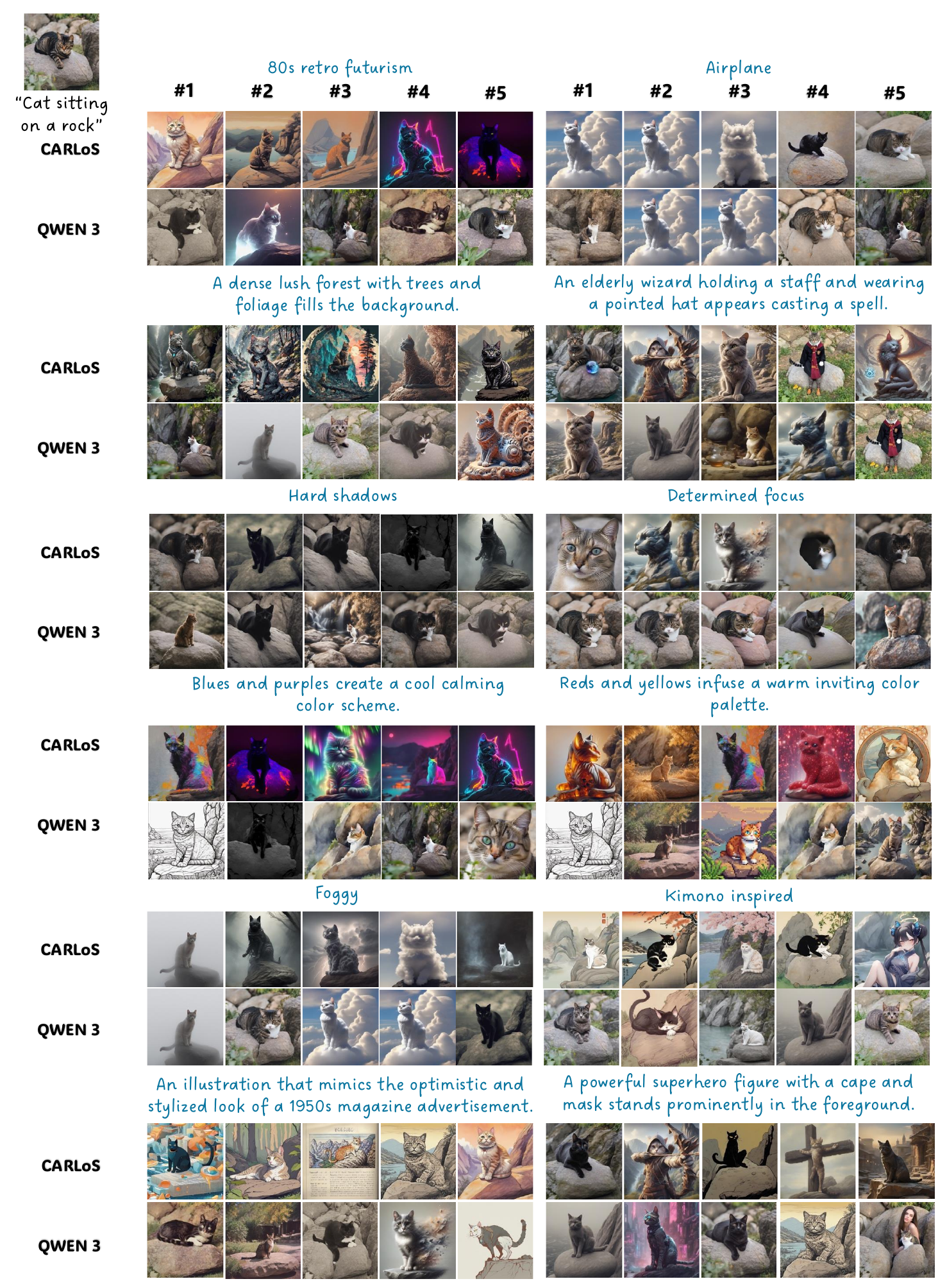}
    \caption{A qualitative comparison showcasing the top-5 LoRAs retrieved by CARLoS (ours, top row) and the Qwen3 textual baseline (bottom row) for 12 uncurated, randomly selected retrieval queries.  The generations all use a fixed base prompt ('Cat sitting on a rock') to better isolate and highlight the stylistic and semantic shifts induced by the retrieved LoRAs.  This figure further demonstrates CARLoS's ability to consistently retrieve LoRAs that are visually and semantically relevant to the query, outperforming the textual retriever, especially for abstract concepts (e.g., '80s retro futurism', 'Kimono inspired', and 'Hard shadows'). Zoomed in viewing recommended.}
    \label{fig:random_comparison}
\end{figure*}
\section{Extended Qualitative Retrieval Comparison}
We provide an extended qualitative evaluation in Figure \ref{fig:random_comparison} to give the reader a granular view of the retrieval performance distribution, comparing our CARLoS retrieval method against our strongest textual baseline, Qwen3. The examples showcase the top-5 retrieved LoRAs for a selection of 12 diverse queries, all applied to a fixed base prompt (e.g., 'A cat sitting on a rock') for a clearer visual comparison of the adapter's effect.

The following analysis summarizes the principal qualitative findings from these examples, emphasizing the patterns that explain CARLoS’s retrieval advantages.

\paragraph{Semantic and Stylistic Precision:} For queries relating to specific artistic or lighting styles, such as '80s retro futurism', 'Blues and purples create a cool calming color scheme', and 'Hard shadows', CARLoS retrieval is highly effective. It consistently returns LoRAs that visually manifest the queried effect across the top-5 ranks. This is a direct result of our approach, which compares the query's semantic shift to the measured generative effect of the LoRAs.
\paragraph{Failure of Textual Baselines:} The textual retriever, Qwen3, often fails to match the query to the correct generative effect, particularly for abstract or complex concepts. For example, for '80s retro futurism', the Qwen3 results show minimal or inconsistent stylistic changes. This suggests that the text retrieval is often misled by irrelevant keywords or unreliable user-provided descriptions, as discussed in our main paper (Section~\ref{sec:qualitative}).
\paragraph{Robustness via Filtering:} The comparative view illustrates the importance of our Strength and Consistency filtering. CARLoS's generations maintain a better adherence to the base prompt because its retrieval process explicitly filters out LoRAs that are too strong (which can override the prompt) or too inconsistent (unpredictable). Conversely, the textual retriever, Qwen3, includes several LoRAs that would be filtered by CARLoS due to low consistency and/or low strength. For example, Qwen3's result \#3 for 'A dense lush forest...' shows a weak effect that fails to consistently apply the style, demonstrating the need to filter out unreliable LoRAs. This filtering ensures the resulting generations are higher quality and more predictable.

 In summary, Figure \ref{fig:random_comparison} provides visual evidence that retrieving LoRAs based on their unbiased, generative effect yields superior and more consistent results, solidifying the findings of our quantitative and subjective evaluations.

\section{Extended Quantitative Evaluation}
\label{sec:suppl_extended_eval}

Table~\ref{tab:supplementary_quantitative_transposed} extends the main results of the main paper (Table~\ref{tab:quantitative}) by reporting retrieval performance for $k = 1 \dots 7$ across all evaluators. Whereas Table~\ref{tab:quantitative} averages evaluator scores over the top 3 retrieved LoRAs, this table expands the analysis to multiple top-$k$ settings. For each column, we average the scores of the top-k ranked retrieved LoRAs. The scores were min-max linearly normalized per-evaluator, and across all queries, top-7 retrieved images, and retrieval methods. This is the exact same normalization mapping as used in Table~\ref{tab:quantitative} for comparability. Specifically, the column \textbf{Top-3} in Table~\ref{tab:supplementary_quantitative_transposed} matches the measurement reported in the main paper in Table~\ref{tab:quantitative}.
The trends confirm that CARLoS consistently surpasses text-based baselines for all $k$, indicating that CARLoS retrieves highly relevant LoRAs along multiple ranking spans, which is a desirable property for practical search, aiming for a stable, user-friendly environment.

\begin{table}[!t]
\centering
\caption{\textbf{Top-7 Retrieval Performance.} Scores indicate the quality of retrieved LoRAs as judged by state-of-the-art Vision-Language Models. CARLoS consistently yields results preferred by all evaluators. The scores are normalized in a min-max manner across all queries and retrievers.}
\label{tab:supplementary_quantitative_transposed}
\resizebox{\columnwidth}{!}{
\begin{tabular}{@{}llccccccc@{}}
\toprule
\textbf{Retriever} & \textbf{Evaluator} & \textbf{Top-1} & \textbf{Top-2} & \textbf{Top-3} & \textbf{Top-4} & \textbf{Top-5} & \textbf{Top-6} & \textbf{Top-7} \\
\midrule
\multirow{4}{*}{E5} & SigLIP2 & 0.317 & 0.297 & 0.289 & 0.279 & 0.271 & 0.267 & 0.263 \\
 & Qwen2.5 & 0.501 & 0.488 & 0.480 & 0.474 & 0.470 & 0.467 & 0.464 \\
 & IR & 0.468 & 0.458 & 0.449 & 0.443 & 0.439 & 0.435 & 0.432 \\
 & HPS & 0.575 & 0.570 & 0.565 & 0.562 & 0.561 & 0.559 & 0.558 \\
\midrule
\multirow{4}{*}{GTE} & SigLIP2 & 0.265 & 0.256 & 0.258 & 0.254 & 0.250 & 0.246 & 0.244 \\
 & Qwen2.5 & 0.470 & 0.461 & 0.461 & 0.457 & 0.454 & 0.452 & 0.451 \\
 & IR & 0.451 & 0.438 & 0.440 & 0.435 & 0.433 & 0.432 & 0.433 \\
 & HPS & 0.562 & 0.554 & 0.556 & 0.554 & 0.553 & 0.553 & 0.553 \\
\midrule
\multirow{4}{*}{BGE} & SigLIP2 & 0.218 & 0.206 & 0.199 & 0.194 & 0.191 & 0.189 & 0.187 \\
 & Qwen2.5 & 0.450 & 0.434 & 0.429 & 0.425 & 0.421 & 0.419 & 0.417 \\
 & IR & 0.408 & 0.390 & 0.387 & 0.387 & 0.384 & 0.381 & 0.378 \\
 & HPS & 0.553 & 0.544 & 0.543 & 0.543 & 0.542 & 0.540 & 0.538 \\
\midrule
\multirow{4}{*}{Qwen3} & SigLIP2 & \underline{0.343} & \underline{0.320} & \underline{0.307} & \underline{0.298} & \underline{0.291} & \underline{0.284} & \underline{0.278} \\
 & Qwen2.5 & \underline{0.521} & \underline{0.505} & \underline{0.495} & \underline{0.488} & \underline{0.484} & \underline{0.480} & \underline{0.477} \\
 & IR & \underline{0.509} & \underline{0.499} & \underline{0.491} & \underline{0.484} & \underline{0.479} & \underline{0.476} & \underline{0.473} \\
 & HPS & \underline{0.599} & \underline{0.593} & \underline{0.590} & \underline{0.587} & \underline{0.585} & \underline{0.583} & \underline{0.582} \\
\midrule
\multirow{4}{*}{\textbf{CARLoS}} & SigLIP2 & \textbf{0.385} & \textbf{0.368} & \textbf{0.350} & \textbf{0.342} & \textbf{0.334} & \textbf{0.328} & \textbf{0.324} \\
 & Qwen2.5 & \textbf{0.561} & \textbf{0.547} & \textbf{0.532} & \textbf{0.524} & \textbf{0.520} & \textbf{0.515} & \textbf{0.512} \\
 & IR & \textbf{0.531} & \textbf{0.520} & \textbf{0.505} & \textbf{0.498} & \textbf{0.493} & \textbf{0.488} & \textbf{0.487} \\
 & HPS & \textbf{0.607} & \textbf{0.603} & \textbf{0.596} & \textbf{0.594} & \textbf{0.591} & \textbf{0.590} & \textbf{0.589} \\
\bottomrule
\end{tabular}
}
\end{table}

\section{Retrieval Diversity and Non-Bias Analysis}

The superior retrieval performance of CARLoS is not solely due to high accuracy but also stems from its ability to select from a \textbf{diverse and broad range} of semantically relevant LoRAs, demonstrating a low bias towards highly popular or overly strong adapters. This section provides both qualitative and quantitative evidence for this crucial property.
\subsection{Qualitative Analysis: Retrieval Frequency}

To confirm that CARLoS does not simply rely on a small set of popular LoRAs, we analyzed the retrieval frequency across our comprehensive benchmark of over 700 queries.

\begin{figure}[t]
    \centering
    \includegraphics[width=\columnwidth]{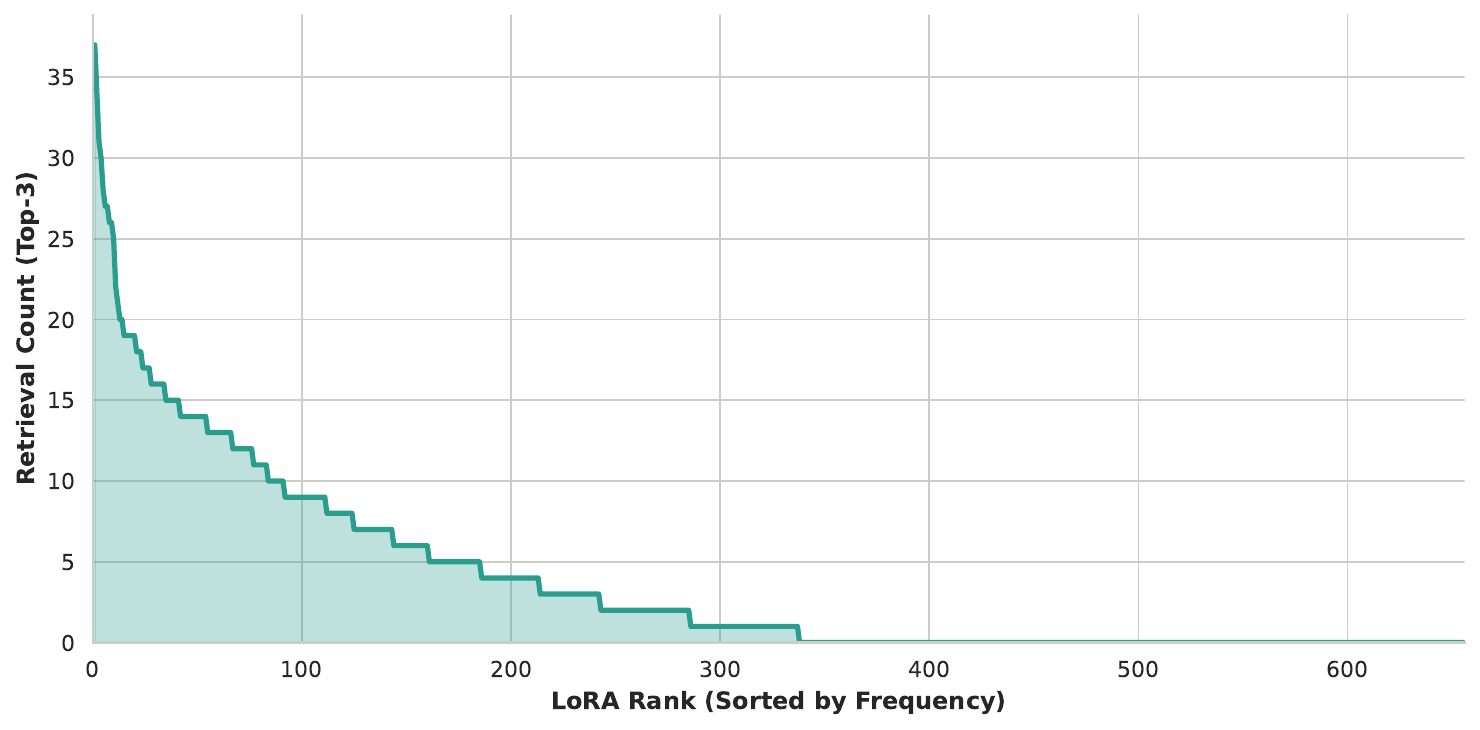}
    \caption{LoRA Retrieval Frequency (Top-3) across the full query set. The graph shows the number of times each unique LoRA appeared in the top-3 results, ordered by frequency, demonstrating that our CARLoS retrieval method utilizes a \textbf{diverse and broad range} of the LoRA corpus, rather than relying on a small, popular subset. The long, shallow tail indicates that the majority of LoRAs in the corpus are retrieved at least once.}
    \label{fig:retrieval_diversity}
\end{figure}

The analysis, summarized in Figure \ref{fig:retrieval_diversity}, plots the number of times each unique LoRA appeared in the top-3 retrieved results, sorted by descending frequency.

\paragraph{Broad Coverage:} The curve exhibits a long, shallow tail, confirming that the retrieval process is highly diverse. Out of the 656 available LoRAs, the majority are retrieved at least once across the benchmark. This diversity is crucial, as it indicates that the system is not strictly biased towards LoRAs with extreme Strength or by social factors, which often plague popularity-based discovery methods.

This analysis complements the qualitative results shown in Figures \ref{fig:teaser}, \ref{fig:gallery}, and \ref{fig:qualitative_comparison_to_text_retrieval} of the main paper, which also depict a variety of effects being retrieved for different queries. The measured diversity underscores the value of our prompt-independent behavioral representation in fostering a non-biased, standardized, and transparent ecosystem for generative components.

\subsection{Quantitative Analysis: Diversity Metrics}
\begin{table}[ht]
\centering
\small
\setlength{\tabcolsep}{3pt}
\begin{tabular}{lccc}
\toprule
\textbf{Retriever} &
\textbf{\begin{tabular}[c]{@{}c@{}}Normalized\\ Entropy $\uparrow$\end{tabular}} &
\textbf{\begin{tabular}[c]{@{}c@{}}Gini\\ Coefficient $\downarrow$\end{tabular}} &
\textbf{\begin{tabular}[c]{@{}c@{}}Effective\\ LoRA Count $\uparrow$\end{tabular}} \\
\midrule
\multicolumn{4}{c}{\textbf{Top-1}}\\
\midrule
E5 & \underline{0.728} & \underline{0.861} & \underline{112.444} \\
GTE & 0.609 & 0.912 & 52.002 \\
BGE & 0.648 & 0.899 & 67.080 \\
Qwen3 & 0.712 & 0.876 & 101.578 \\
CARLoS & \textbf{0.790} & \textbf{0.802} & \textbf{167.901} \\
\midrule
\multicolumn{4}{c}{\textbf{Top-2}}\\
\midrule
E5 & \underline{0.765} & \underline{0.827} & \underline{142.817} \\
GTE & 0.653 & 0.881 & 69.162 \\
BGE & 0.663 & 0.887 & 73.585 \\
Qwen3 & 0.738 & 0.854 & 119.793 \\
CARLoS & \textbf{0.824} & \textbf{0.755} & \textbf{210.096} \\
\midrule
\multicolumn{4}{c}{\textbf{Top-3}}\\
\midrule
E5 & \underline{0.780} & \underline{0.812} & \underline{157.069} \\
GTE & 0.681 & 0.866 & 82.915 \\
BGE & 0.670 & 0.881 & 77.228 \\
Qwen3 & 0.752 & 0.842 & 130.908 \\
CARLoS & \textbf{0.843} & \textbf{0.723} & \textbf{237.598} \\
\midrule
\multicolumn{4}{c}{\textbf{Top-4}}\\
\midrule
E5 & \underline{0.792} & \underline{0.796} & \underline{170.462} \\
GTE & 0.704 & 0.851 & 96.214 \\
BGE & 0.676 & 0.877 & 80.008 \\
Qwen3 & 0.758 & 0.837 & 136.115 \\
CARLoS & \textbf{0.852} & \textbf{0.708} & \textbf{250.694} \\
\midrule
\multicolumn{4}{c}{\textbf{Top-5}}\\
\midrule
E5 & \underline{0.798} & \underline{0.788} & \underline{176.911} \\
GTE & 0.721 & 0.839 & 107.057 \\
BGE & 0.677 & 0.874 & 80.632 \\
Qwen3 & 0.761 & 0.833 & 139.514 \\
CARLoS & \textbf{0.859} & \textbf{0.693} & \textbf{263.603} \\
\midrule
\multicolumn{4}{c}{\textbf{Top-6}}\\
\midrule
E5 & \underline{0.803} & \underline{0.781} & \underline{183.273} \\
GTE & 0.735 & 0.827 & 117.830 \\
BGE & 0.678 & 0.873 & 81.509 \\
Qwen3 & 0.765 & 0.830 & 142.714 \\
CARLoS & \textbf{0.865} & \textbf{0.682} & \textbf{273.499} \\
\midrule
\multicolumn{4}{c}{\textbf{Top-7}}\\
\midrule
E5 & \underline{0.808} & \underline{0.775} & \underline{188.747} \\
GTE & 0.747 & 0.818 & 126.985 \\
BGE & 0.681 & 0.870 & 82.737 \\
Qwen3 & 0.765 & 0.829 & 143.187 \\
CARLoS & \textbf{0.869} & \textbf{0.672} & \textbf{281.016} \\
\bottomrule
\end{tabular}
\caption{Retrieval distribution metrics across retrievers for each Top-$k$. Higher is better for Normalized Entropy and Effective LoRA Count; lower is better for Gini Coefficient. \textbf{Bold} marks best; \underline{underline} second best.}
\vspace{-0.5cm}
\label{tab:retrieval_metrics_narrow}
\end{table}

To quantitatively demonstrate that our method relies on semantic relevance rather than just a small set of "popular" adapters, we rigorously assess the diversity and non-bias of our CARLoS retrieval method, by reporting three complementary measures of distributional diversity (Table~\ref{tab:retrieval_metrics_narrow}): Normalized Entropy, Gini Coefficient, and Effective Count. Entropy captures overall uncertainty and sensitivity to rare outcomes, the Gini Coefficient quantifies inequality and concentration of mass, and the Effective Count translates these abstract measures into an interpretable “number of active components”. Together, they provide a robust and multi-perspective characterization of the distribution’s spread and skewness.

Table~\ref{tab:retrieval_metrics_narrow} shows that CARLoS consistently achieves the highest $\text{Normalized Entropy}$ and $\text{ELC}$ and the lowest $\text{Gini Coefficient}$ across all ranking spans ($k=1$ to $k=7$). Moreover, looking at Qwen3, the next best retrieval method after CARLoS as evident in Table~\ref{tab:supplementary_quantitative_transposed}, seems to fall behind in all diversity measures. This collective evidence asserts the superior performance of CARLoS not only in measures of accuracy but also through genuine and widespread semantic matches across the LoRA corpus, demonstrating the non-biased utility of our behavioral representation.

\section{Analysis of Strength and LoRA Scale}
\label{sec:strength_vs_scale}

\begin{figure*}[!t]
    \centering
    \includegraphics[
        width=0.8\textwidth
    ]{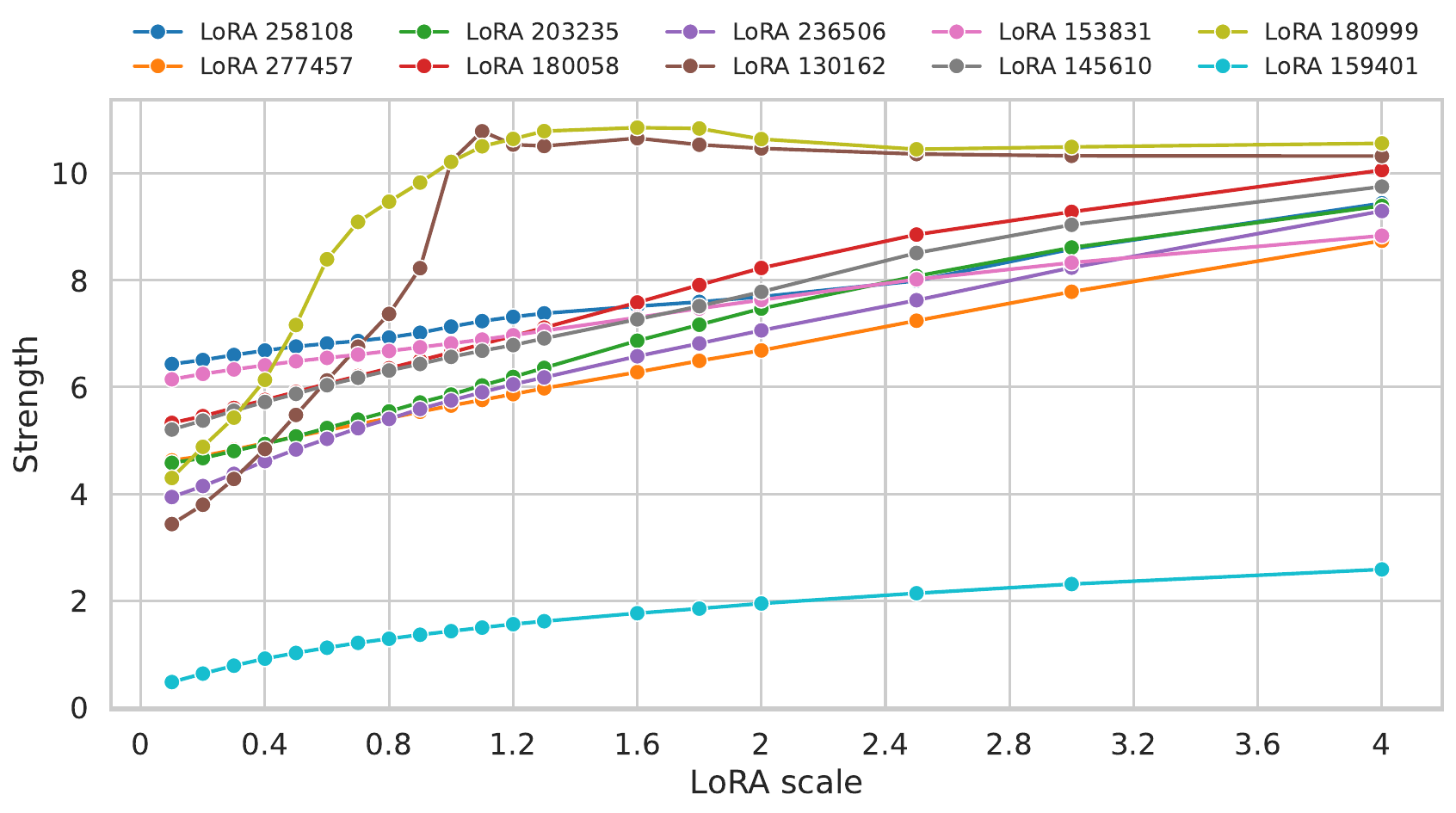}
    \caption{Strength vs. LoRA Scale, for 10 distinct LoRAs selected to represent a range of behaviors in the corpus. The plot shows that while Strength generally increases with scale, the relationship is not strictly linear and highly variable across different LoRAs.}
    \label{fig:strength_vs_scale} 
\end{figure*}

To quantitatively evaluate the functional relationship between the "LoRA Scale" hyperparameter and our Strength metric, we repeated the indexing and metric calculation process (described in Section \ref{sec:method} of the main paper). This analysis was performed on 10 different LoRAs across a range of different scale values, as depicted in Figure \ref{fig:strength_vs_scale}. To showcase existing behaviors, while maintaining a reasonable computation effort, we purposefully sampled 10 LoRAs spanning low/medium/high Strength and Consistency, thus results are illustrative rather than population-level.

While a positive, monotonically increasing connection is generally visible, the results clearly demonstrate that this relationship is complex and not uniform across different LoRAs. We highlight the following key observations:

\begin{itemize}
    \item \textbf{Saturation and Diminishing Returns:} The Strength of some LoRAs (e.g., 180999 and 130162) appears to saturate at a certain scale. Beyond this point, the Strength plateaus or, in some cases, even softly decreases, indicating a non-linear response.
    
    \item \textbf{Non-Uniform Trajectories:} The functional relationship is not strictly linear. Different LoRAs exhibit distinct initial "biases" (i.e., base Strength at low scales, such as LoRA 159401) and different "incline ratios" or slopes. This variation can even cause trajectories to cross (e.g., LoRAs 153831 and 180058).
\end{itemize}

These findings suggest that, due to the observed variability, relying on the LoRA scale setting alone to accurately predict a LoRA's full Strength-to-scale trajectory is unreliable across the ecosystem. Furthermore, extrapolating from a LoRA's Strength at a single, arbitrary scale is unreliable.

Therefore, this analysis confirms that the CARLoS Strength metric (which we calculate at a fixed scale of 1.0) captures intrinsic, valuable information about a LoRA's behavior. This information is distinct from, and cannot be acquired solely by, observing the LoRA scale hyperparameter. 

\section{Implementation and Curation Details}
\label{sec:suppl_implementation_details}
We provide implementation details to facilitate reproducibility and future extensions.
\subsection{Prompt Set Construction}
\label{sec:suppl_prompt_construction}

The prompt sets used for indexing and retrieval, as described in Section 3.1, were constructed using ChatGPT 4o. We guided the LLM generation process with the following sequence of prompts to ensure alignment with common community usage on CivitAI:

\begin{enumerate}
    \item \textit{what are the most popular categories for prompts in the context of text to image ai generations?}
    \item \textit{Regarding each category, for each subcategory, provide me with 16 prompt. make sure these prompts are as varied as possible per-category, and that they do not exceed the length of 73 CLIP tokens. keep the semantics safe for work.}
\end{enumerate}

This process yielded a total of 560 unique prompts. This set was then evenly divided into two disjoint subsets:
\begin{itemize}
    \item \textbf{Indexing Set ($\mathcal{P}$):} $N=280$ prompts, used for generating images to index the LoRA corpus (Section 3.2).
    \item \textbf{Retrieval Set ($\mathcal{P'}$):} $N=280$ prompts, used as the base for creating the reciprocal textual CLIP-diff query vector (Section 3.3).
\end{itemize}

$\mathcal{P}$ and $\mathcal{P'}$ are disjoint and match the same category/sub-category taxonomy. Prompts were assigned deterministically (first 8 to $\mathcal{P}$, last 8 to $\mathcal{P'}$) Example prompts from both sets, along with their associated categories and subcategories, are provided in Table~\ref{tab:prompts_indexing_sample} and Table~\ref{tab:prompts_retrieval_sample}. The complete set of 560 prompts is included in the supplementary materials as \texttt{indexing\_and\_retrieval\_prompts.py} to ensure reproducibility.

\begin{table*}[!t]
\centering
\caption{A sampled subset of prompts from indexing set $\mathcal{P}$ (full lists in code).}
\label{tab:prompts_indexing_sample}
\resizebox{\textwidth}{!}{
\footnotesize
\begin{tabular}{@{}p{0.17\linewidth} p{0.27\linewidth} p{0.53\linewidth}@{}}
\toprule
\textbf{Category} & \textbf{Sub-Category} & \textbf{Example Prompt} \\
\midrule
Animals & Dragons, Unicorns, Phoenixes & A colossal fire-breathing dragon perched atop a castle ruin \\
Animals & Hybrid Creatures & A cat-mermaid lounging on a rock, tail shimmering in the sunlight \\
Animals & Realistic Wildlife & A majestic snow leopard walking across rocky terrain, snow flurries in the air \\
Artistic\_Styles & Cyberpunk, Synthwave, Vaporwave & A neon-drenched cyberpunk alleyway with holographic signs and flying cars \\
Artistic\_Styles & Oil Painting, Watercolor, Digital Painting & A Baroque-style oil painting of a regal king in ornate armor \\
Cinematic & Dynamic Action Shots & A warrior leaping through the air, sword drawn, against a fiery backdrop \\
Conceptual\_Arts & Dreamlike or Mind-Bending Scenes & A floating island with a waterfall that cascades into the sky \\
Fashion & Couture and Avant-Garde & A runway model wearing an avant-garde dress made of reflective shards \\
Landscapes & Natural Landscapes & A breathtaking sunrise over a vast mountain range, golden mist in the valleys \\
Portraits & Fantasy and Sci-Fi Characters & A regal elf queen with silver hair, glowing blue eyes, intricate gold armor \\
Vehicles & Steampunk Machinery and Robots & A massive steampunk airship floating above a Victorian city \\
\bottomrule
\end{tabular}}
\end{table*}

\begin{table*}[!t]
\centering
\caption{A sampled subset of prompts from retrieval set $\mathcal{P}'$ (constructed independently of $\mathcal{P}$).}
\label{tab:prompts_retrieval_sample}
\resizebox{\textwidth}{!}{
\footnotesize
\begin{tabular}{@{}p{0.17\linewidth} p{0.27\linewidth} p{0.53\linewidth}@{}}
\toprule
\textbf{Category} & \textbf{Sub-Category} & \textbf{Example Prompt} \\
\midrule
Animals & Dragons, Unicorns, Phoenixes & A unicorn with a mane of galaxies galloping through space \\
Animals & Realistic Wildlife & A red fox standing alert in a snowy forest, breath visible in the cold air \\
Artistic\_Styles & Cyberpunk, Synthwave, Vaporwave & A vaporwave city square with glitching sculptures and retro grids \\
Artistic\_Styles & Oil Painting, Watercolor, Digital Painting & A luminous digital painting of a cathedral interior with stained glass glow \\
Cinematic & Scenes That Look Like Movie Frames & A dimly lit interrogation room with a single overhead lamp \\
Conceptual\_Arts & Dreamlike or Mind-Bending Scenes & A corridor of doors opening into different skies and seasons \\
Fashion & Couture and Avant-Garde & A gown constructed from holographic fabric, refracting rainbow light \\
Landscapes & Sci-Fi and Futuristic Worlds & A floating island city powered by advanced wind turbines, hovering over the ocean \\
Portraits & Fantasy and Sci-Fi Characters & A biomechanical cyborg with glowing circuits and synthetic skin \\
Vehicles & Futuristic Cars, Bikes, Spaceships & A modular space shuttle docking with a rotating ring station \\
\bottomrule
\end{tabular}}
\end{table*}

\subsection{LoRA Corpus Curation}
\label{sec:suppl_lora_corpus}

Our LoRA corpus $\mathcal{C}$ was collected from CivitAI, the largest public repository for such models, using their public API. The curation process involved several filtering and validation steps:

\begin{itemize}
    \item We initially retrieved the metadata for the first 10,000 reachable SDXL LoRAs via the CivitAI API. The API pagination used default ordering, in practice this correlates with downloads.
    \item This set was filtered to exclude modules that were (a) less than 100 days old, to avoid transient or unstable uploads, and (b) had a file size exceeding 10 GB, which implies a wrong tagging as LoRA, since typical LoRA sizes are much smaller.
    \item To prioritize testing more stable models, we focused our downloading efforts on the top 1,875 most popular LoRAs from the filtered set, ordered by their 'downloads' attribute.
    \item We then proceeded to download the weights file for each LoRA, skipping any with missing files.
    \item Finally, each downloaded LoRA was programmatically validated by attempting to load it into a standard \texttt{diffusers} SDXL pipeline.
\end{itemize}

From an initial pool of 1,875 filtered metadata entries, we successfully validated and loaded 656 LoRAs. This validation success rate ($\frac{656}{1875}\approx35\%$) highlights the significant portion of non-functional, corrupted, or otherwise unusable modules present in public repositories, underscoring the challenge addressed by our curation process. This final, validated set of 656 LoRAs constitutes our indexed corpus. A detailed list of all indexed LoRAs - containing their names and model IDs as they appear in CivitAI - is attached in a file named \texttt{lora\_names\_with\_civit\_ids.csv}.

\subsection{LoRA Indexing and Embedding}
\label{sec:suppl_indexing}

The LoRA indexing process, detailed in Section 3.2, was implemented using the \texttt{diffusers} Python library \cite{von2022diffusers}.

\paragraph{Image Generation.}
We used the \texttt{StableDiffusionXLPipeline} module with the \texttt{stable-diffusion-xl-base-1.0} base model. To enforce uniform conditions and enable parallel processing, each of the 656 LoRAs was indexed in an independent process.

For each LoRA, the process first loaded the base SDXL model and then applied the LoRA weights. We generated $M=16$ images for each of the $N=280$ indexing prompts ($\mathcal{P}$). The generation process began with a fixed random seed of $42$, which was sequentially incremented for each of the $M=16$ samples. We used a fixed \textbf{LoRA scale of 1.0} for all generations, leaving other hyperparameters at their defaults (e.g., $CFG=7.5$, Euler Scheduler). A separate, one-time process was run to generate the vanilla (no-LoRA) images for all (prompt, seed) pairs.

\paragraph{CLIP Embedding.}
Following generation, we embedded all images using the \texttt{transformers} library \cite{wolf2020transformers}. We employed the \texttt{CLIPModel} with the \texttt{openai/clip-vit-base-patch32} variant. The \textbf{non-normalized, 512-dimensional} CLIP-space vectors were extracted using the \texttt{get\_image\_features} function.

\paragraph{Storage Optimization}
To optimize storage requirements while preserving sufficient data for future reconstruction validation, the original 1024x1024 generated images were replaced with 64x64 thumbnails (approximately 10 KB each). Importantly, this step was performed after the CLIP embedding vectors were computed and did not impact our resulting metrics. The 512-dimensional embedding vectors were saved as raw data (approximately 4 KB each). The corpus of CLIP-diff vectors—derived by subtracting the vanilla embedding from the LoRA-modified embedding for each (prompt, seed) pair—forms the core raw data for our analysis.

\section{User Study Additional Details}
\subsection{Study Methodology and Interface}
To validate our quantitative Vision-Language Model (VLM) results with human judgment, we conducted a double-blind subjective user study involving 36 unique participants. The study was deployed using Google Forms. For each comparison, participants were shown the results from two retrieval methods: our proposed CARLoS method and one of the four textual baselines (Qwen3, E5, GTE, or BGE).

\begin{figure}[ht]
    \centering
    \includegraphics[width=0.95\columnwidth]{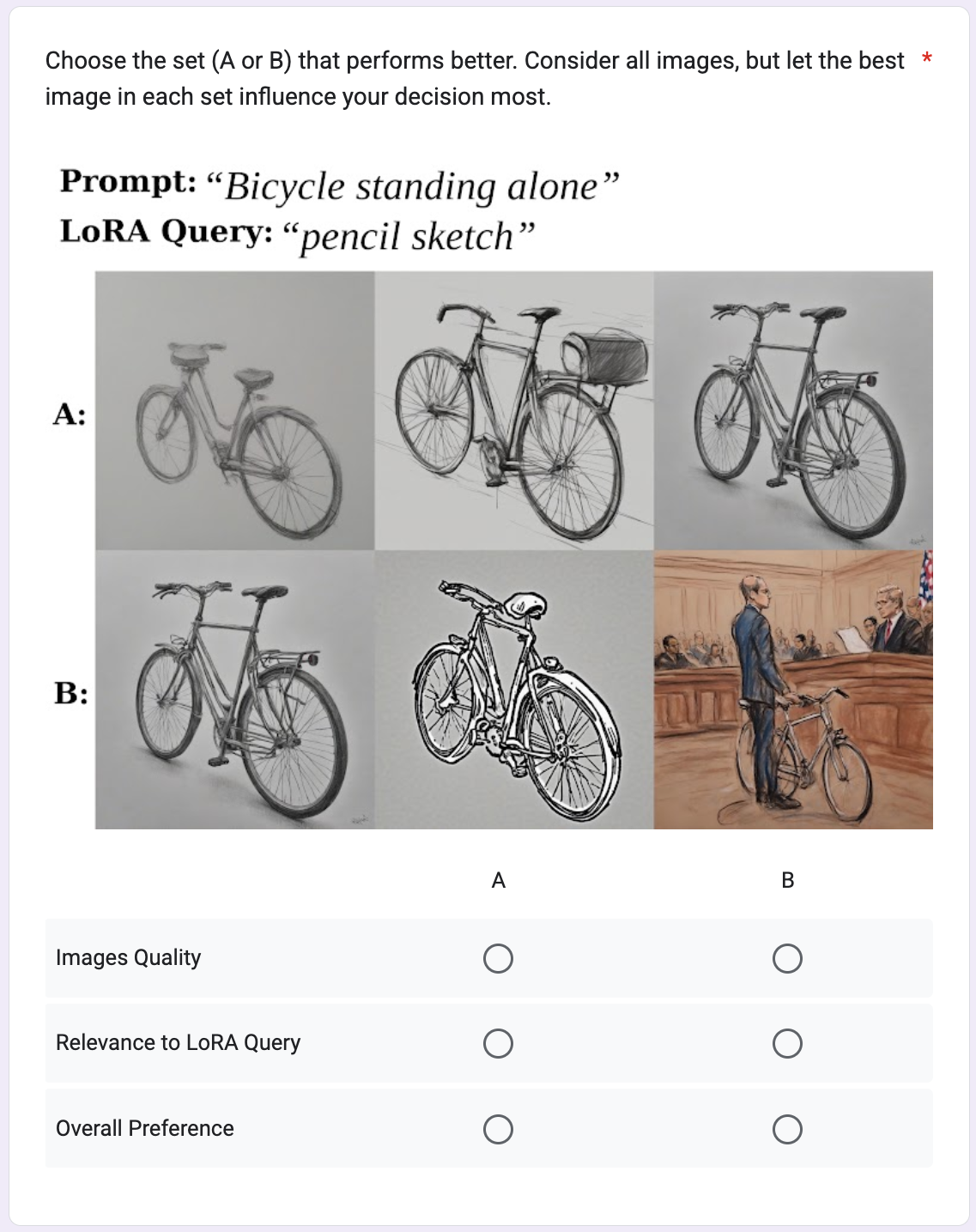}
    \caption{A representative screenshot of the interface used for the double-blind subjective user study. Each question presented two sets of images, labeled 'A' and 'B', generated by the top-3 LoRAs retrieved by CARLoS and a textual baseline for a specific query (e.g., ``pencil sketch''). Participants evaluated the sets based on Image Quality, Relevance to the LoRA Query, and Overall Preference. Note that the best image in each set was emphasized as a deciding factor.}
    \label{fig:user_study_screenshot}
\end{figure}

\subsubsection{Procedure and Randomization}
To ensure a rigorous and unbiased evaluation, we implemented the following randomization protocols:
\begin{itemize}
    \item \textbf{Query Selection:} Approximately 100 unique text queries were randomly selected from our comprehensive benchmark set of over 700 queries.
    \item \textbf{Set Labeling:} The results from the CARLoS method were randomly assigned to either "Set A" or "Set B" to maintain the double-blind nature of the study, preventing participants from knowing which set was ours.
    \item \textbf{Baseline Pairing:} CARLoS was randomly paired against one of the four textual retrieval baselines (Qwen3, E5, GTE, BGE) for each question.
    \item \textbf{Image Presentation:} Both sets, A and B, displayed images generated by the respective method's top-3 retrieved LoRAs, applied to a fixed base prompt (e.g., ``Bicycle standing alone'').
\end{itemize}

\subsubsection{Evaluation Metrics and Aggregation}
For each comparison, participants were instructed to evaluate the two image sets based on three criteria (see Figure \ref{fig:user_study_screenshot}):
\begin{enumerate}
    \item \textbf{Images Quality:} Aesthetic and technical quality of the generations.
    \item \textbf{Relevance to LoRA Query:} How accurately the overall set reflects the intended semantic or stylistic shift requested by the query.
    \item \textbf{Overall Preference:} The user's final, holistic choice.
\end{enumerate}
Participants chose between Set A and Set B, for each of the three metrics. A total of approximately 300 unique comparison sub-questions (100 questions $\times$ 3 metrics) were generated. Each sub-question was answered by a minimum of 6 participants, leading to the aggregated preference scores of more than 1800 answered sub-questions reported in Figure~\ref{fig:user_study} of the main paper.

\subsection{Participant Details}
The study involved 36 unique individual human responders. The participants were recruited from a varied pool, spanning different age groups, genders, and professional occupations. All participants were well-informed of the purpose of the study, which was to assess the quality of generative AI components. This demographic diversity helps ensure that the aggregated results reflect a broad user preference.

\end{document}